\documentclass[letterpaper, 10 pt, conference]{ieeeconf}  

\IEEEoverridecommandlockouts                              

\usepackage{graphicx} 
\usepackage{amsmath} 
\usepackage{amsfonts} 
\usepackage{booktabs,multirow} 
\usepackage{algpseudocode} 
\usepackage[ruled,linesnumbered]{algorithm2e} 
    
\usepackage{bm}
\usepackage{siunitx}
\usepackage{pgf}
\usepackage{tikz}
\usetikzlibrary{bayesnet}
\usetikzlibrary{positioning}
\usetikzlibrary{calc}
\usepackage{wasysym}
\usepackage{makecell}
\usepackage{amssymb}
\usepackage[pdfborder={0 0 0.1}]{hyperref}
\usepackage[capitalise]{cleveref}

\definecolor{mpl_propcycle_1}{HTML}{1f77b4}
\definecolor{mpl_propcycle_2}{HTML}{ff7f0e}
\definecolor{mpl_propcycle_3}{HTML}{2ca02c}
\definecolor{mpl_propcycle_4}{HTML}{d62728}

\def\sC{{\mathcal{C}}}

\def\dN{{\mathcal{N}}}

\def\vx{{\bm{x}}}
\def\vu{{\bm{u}}}
\def\vz{{\bm{z}}}
\def\valpha{{\bm{\alpha}}}
\def\vbeta{{\bm{\beta}}}

\def\veps{{\bm{\epsilon}}}

\def\mQ{{\bm{Q}}}

\def\lD{{\mathrm{D}}}

\def\cpx{{p_\mathrm{x}}}
\def\cpy{{p_\mathrm{y}}}
\def\qctx{q_{\mathrm{ctx}}}
\def\qfwd{q_{\mathrm{fwd}}}
\def\aterrfcn{\alpha_{\mathrm{terrain}}}
\def\tterrain{\tau_{\mathrm{terrain}}}
\def\arobot{\valpha_{\mathrm{robot}}}

\SetKwInput{KwInput}{Input}

\title{\LARGE \textbf{Context-Conditional Navigation with a Learning-Based Terrain- and Robot-Aware Dynamics Model}
}

\author{Suresh Guttikonda$\,{}^{*,1,2}$, Jan Achterhold$\,{}^{*,1}$, Haolong Li$\,{}^{1}$, Joschka Boedecker$\,{}^{2}$, and Joerg Stueckler$\,{}^{1}$
\thanks{\hspace{-1.5em} This work has been supported by Max Planck Society and Cyber Valley. The authors thank the International
Max Planck Research School for Intelligent Systems (IMPRS-IS) for supporting Jan Achterhold and Haolong Li.}
\thanks{\hspace{-1.5em} ${}^{1}$Embodied Vision Group, Max Planck Institute for Intelligent Systems, Tuebingen, Germany, ${}^{2}$University of Freiburg, Germany}%
\thanks{\hspace{-1.5em} ${}^{*}$equal contribution\vspace*{0.3em}} %
\thanks{\hspace{-1em}{Source code for our approach is available at \url{https://github.com/EmbodiedVision/tradyn}.}}%
}

\newcommand\copyrighttext{%
  \footnotesize \textcopyright 2023 IEEE. Personal use of this material is permitted.
  Permission from IEEE must be obtained for all other uses, in any current or future 
  media, including reprinting/republishing this material for advertising or promotional 
  purposes, creating new collective works, for resale or redistribution to servers or 
  lists, or reuse of any copyrighted component of this work in other works. 
  }
\newcommand\copyrightnotice{%
\begin{tikzpicture}[remember picture,overlay]
\node[anchor=south,yshift=10pt] at (current page.south) {\fbox{\parbox{\dimexpr\textwidth-\fboxsep-\fboxrule\relax}{\copyrighttext}}};
\end{tikzpicture}%
}

\newcommand\acceptancetext{%
  \footnotesize Accepted for publication in European Conference on Mobile Robots (ECMR), 2023. Version including corrections (see  p. 8).
  }
\newcommand\acceptancenotice{%
\begin{tikzpicture}[remember picture,overlay]
\node[anchor=north,yshift=-20pt] at (current page.north)
{\parbox{\dimexpr\textwidth-\fboxsep-\fboxrule\relax}
{\acceptancetext}};
\end{tikzpicture}%
}

\begin{document}

\maketitle
\acceptancenotice
\copyrightnotice
\thispagestyle{empty}
\pagestyle{empty}

\begin{abstract}
In autonomous navigation settings, several quantities can be subject to variations. Terrain properties such as friction coefficients may vary over time depending on the location of the robot. Also, the dynamics of the robot may change due to, e.g., different payloads, changing the system's mass, or wear and tear, changing actuator gains or joint friction.
An autonomous agent should thus be able to adapt to such variations.
In this paper, we develop a novel probabilistic, \underline{t}errain- and \underline{r}obot-\underline{a}ware forward \underline{\smash{dyn}}amics model, termed TRADYN, which is able to adapt to the above-mentioned variations. It builds on recent advances in meta-learning forward dynamics models based on Neural Processes.
We evaluate our method in a simulated 2D navigation setting with a unicycle-like robot and different terrain layouts with spatially varying friction coefficients.
In our experiments, the proposed model exhibits lower prediction error for the task of long-horizon trajectory prediction, compared to non-adaptive ablation models. 
We also evaluate our model on the downstream task of navigation planning, which demonstrates improved performance in planning control-efficient paths by taking robot and terrain properties into account.

\end{abstract}

\section{INTRODUCTION}
Autonomous mobile robot navigation-- the robot's ability to reach a specific goal location -- has been an attractive research field over several decades, with applications ranging from self-driving cars, warehouse and service robots, to space robotics. 
In certain situations, e.g. weeding in agricultural robotics or search and rescue operations, robots operate in harsh and unstructured outdoor environments with limited or no human supervision to complete their task. 
During such missions, the robot needs to navigate over a wide variety of terrains with changing types, such as grass, gravel, or mud with varying slope, friction, and other characteristics. 
These properties are often hard to fully and accurately model beforehand~\cite{sonker2021adding}.
Moreover, properties of the robot itself can change during operation due to battery consumption, weight changes, or wear and tear of the robot. 
Thus, the robot needs to be able to adapt to both changes in robot-specific and terrain-specific properties.

\begin{figure}[thpb]
    \centering
    \includegraphics[scale=1]{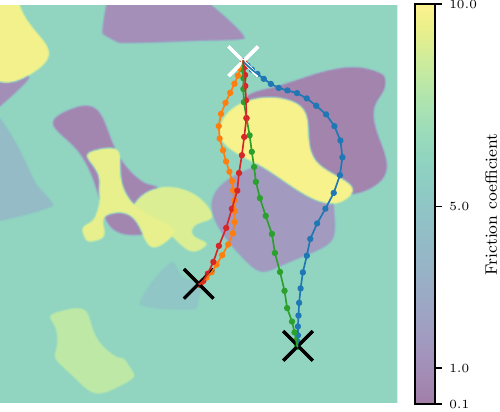}
    \caption{\textbf{Terrain- and robot-aware control-efficient navigation}. We propose a method for control-cost optimal navigation with learned dynamics models. Our method can adapt to varying, unobserved  properties of the robot, such as the mass, and spatially varying properties of the terrain, such as the friction coefficient. In the above example of navigating from a single starting point (white cross) to two different goals (black cross), as a result, our method circumvents areas of high friction coefficient and favors areas of low-friction coefficient. As the dissipated energy also depends on the mass of the robot, a heavy robot ($m= \SI{4}{\kilo\gram}$, {\color{mpl_propcycle_1}blue}, {\color{mpl_propcycle_2}orange}) is allowed to take longer detours to the goal than a light robot ($m= \SI{1}{\kilo\gram}$, {\color{mpl_propcycle_3}green}, {\color{mpl_propcycle_4}red}).}
    \label{fig:tradyn:terrain_aware_motion_planning}
    \vspace{-1em}
\end{figure}

In this work, we develop a novel context-conditional learning approach which captures robot-specific and terrain-specific properties from interaction experience and environment maps.
The idea for adaptability to varying robot-specific properties is to learn a deep forward dynamics model which is conditioned on a latent context variable.
The context variable is inferred online from observed state transitions.
The terrain features are extracted from an environment map and additionally included as conditional variable for the dynamics model.

We develop and evaluate our approach in a 2D simulation of a mobile robot modeled as a point mass with unicycle driving dynamics that depend on a couple of robot-specific and terrain-specific parameters.
Terrains are defined by regions in the map with varying properties. 
We demonstrate that our context-aware dynamics model learning approach can capture the varying robot and terrain properties well, while a dynamics model without context-awareness achieves less accurate prediction and planning performance.

In summary, in this paper, we contribute the following:
\begin{enumerate}
    \item We propose a probabilistic deep forward dynamics model which can adapt to robot- and terrain-specific properties that influence the mobile robot's dynamics. 
    \item We demonstrate in a 2D simulation environment that these adaptation capabilities are crucial for the predictive performance of the dynamics model.
    \item The learned context-aware dynamics model is used for robot navigation using model-predictive control. This way, efficient paths can be planned that take robot and terrain properties into account (see \cref{fig:tradyn:terrain_aware_motion_planning}).
\end{enumerate}

\section{RELATED WORK}

Some approaches to terrain-aware navigation use semantic segmentation for determining the category of terrain and use this information to only navigate on segments of traversable terrain~\cite{valada2017adapnet,yang1028unifying}.
Zhu et al.~\cite{zhu2019offroad} propose to use inverse reinforcement learning to learn the control costs associated with traversing terrain from human expert demonstrations. 
These methods, however, do not learn the dynamical properties of the robot on the terrain classes explicitly like our methods.

In BADGR~\cite{kahn2021badgr} a predictive model is learned of future events based on the current RGB image and control actions, which can be used for planning navigation trajectories.
The predicted events are collision, bumpiness, and position.
The model is trained from sample trajectories in which the events are automatically labelled.
Grigorescu et al.~\cite{grigorescu2021lvdnmpc} learn a vision-based dynamics model which encodes camera images into a state observation for model-predictive control. Different to our approach, however, the method does not learn a model that can capture a variety of terrain- and robot-specific properties jointly.
Siva et al.~\cite{siva2021enhancing} learn an offset model from the predicted to the actual behavior of the robot from multimodal terrain features determined from camera, LiDAR, and IMU measurements. 
In Xiao et al.~\cite{xiao2021learning} a method for learning an inverse kinodynamics model from inertial measurements is proposed to handle high-speed motion planning on unstructured terrain.
Sikand et al.~\cite{sikand2022visual} use contrastive learning to embed visual features of terrain with similar traversability properties close in the feature space.
The terrain features are used for learning preference-aware path planning.
Different to our study, the above approaches do not distinguish terrain- and robot-specific properties and model them concurrently.

Several approaches for learning action-conditional dynamics models have been proposed in the machine learning and robotics literature in recent years.
In the seminal work PILCO~\cite{deisenroth2011pilco}, Gaussian processes are used to learn to predict subsequent states, conditioned on actions.
The approach is demonstrated for balancing and swinging up a cart-pole.
Several approaches learn latent embeddings of images and predict future latent states conditioned on actions using recurrent neural networks~\cite{lenz2015deepmpc,oh2015action,finn2016unsupervised,hafner2019learning}.
The models are used in several of these works for model-predictive control and planning.
Learning-based dynamics models are also popular in model-based reinforcement learning (see e.g.~\cite{nagabandi2018neural}).
Shaj et al.~\cite{shaj2020acrnn} propose action-conditional recurrent Kalman networks which implement observation and action-conditional state-transition models in a Kalman filter with neural networks.
While these approaches can model context from past observations in the latent state of the recurrent neural network, some approaches allow for incorporating an arbitrary set of context observations to infer a context variable~\cite{lee2020context} or a probability distribution thereon~\cite{achterhold2021explore}. 
In this paper, we base our approach on the context-conditional dynamics model learning approach in~\cite{achterhold2021explore} to infer the distribution of a context variable of robot-specific parameters using Neural Processes~\cite{garnelo2018neural}.
%
\section{BACKGROUND}

We build our approach on the context-conditional probabilistic neural dynamics model of \emph{Explore the Context} (EtC~\cite{achterhold2021explore}).
In EtC, the basic assumption is that the dynamical system can be formulated by a Markovian discrete-time state-space model
\begin{equation}
    \label{eq:tradyn:etc_assumption}
    \vx_{n+1} = f( \vx_{n}, \vu_{n}, \valpha ) + \veps_n,~\veps_n \sim \dN( 0, \mQ_n ),
\end{equation}
where~$\vx_{n}$ is the state at timestep~$n$,~$\vu_{n}$ is the control input, and~$\valpha$ is a latent, \emph{unobserved} variable which modulates the dynamics, e.g., robot or terrain parameters. Gaussian additive noise is modeled by $\veps_n$, having a diagonal covariance matrix $\bm{Q}_n$.
Not only $\valpha$ is assumed to be unknown, but also the function $f$ itself. To model the system dynamics, EtC thus introduces an approximate forward dynamics model $\qfwd$. To capture the environment-specific properties $\valpha$, the learned dynamics model is conditioned on a latent context variable~$\vbeta \in \mathbb{R}^B$. A probability density on $\bm{\beta}$ is inferred from interaction experience on the environment, represented by $K$ transitions ($\vx_+ \leftarrow \vx,\vu$) following \cref{eq:tradyn:etc_assumption} and collected in a \emph{context} set 
$
\sC^\alpha = {\{(\vx^{(k)}, \vu^{(k)}, \vx^{(k)}_+)\}}_{k=1}^K.
$ 
A learned \emph{context encoder} $\qctx(\vbeta \mid \sC^\alpha)$ infers the density on $\vbeta$. The target rollout $\lD^\alpha = [\vx_0, \vu_0, \vx_1, \vu_1, \ldots, \vu_{N-1}, \vx_{N}]$ is a trajectory on the environment. Both context set and target rollout are generated on the same environment instance $\bm{\alpha}$.
For a pair of target rollout and context set, the learning objective is to maximize the marginal log-likelihood 
\begin{equation}
   \label{eq:tradyn:mll}
   \log p( \lD^\alpha \mid \sC^\alpha ) = \log \int p( \lD^\alpha \mid \vbeta ) \, p( \vbeta \mid \sC^\alpha ) \, d\vbeta.
\end{equation}
Overall, we aim to maximize $\log p( \lD^\alpha \mid \sC^\alpha )$ in expectation over the distribution of environments $\Omega_\alpha$, and a distribution of pairs of target rollouts and context sets $\Omega_{\lD^\alpha,\sC^\alpha}$, i.e.
\begin{equation}
    \mathbb{E}_{\valpha \sim \Omega_\alpha, (\lD^\alpha,\sC^\alpha) \sim \Omega_{\lD^\alpha,\sC^\alpha}} \left[ \log p( \lD^\alpha \mid \sC^\alpha ) \right].
\end{equation}
The term $p( \lD^\alpha \mid \vbeta )$ is modeled by single-step and multi-step prediction factors and reconstruction factors, all implemented by the approximate dynamics model $\qfwd \left( \vx^{}_{n} \mid \vx^{}_{0}, \vu^{}_{0:n-1}, \vbeta \right)$, while $p(\vbeta \mid \sC^\alpha)$ is approximated by $\qctx(\vbeta \mid \sC^\alpha)$.

Technically, the forward dynamics model is implemented with gated recurrent units (GRU,~\cite{cho2014learning}) in a latent space.
The initial state $\bm{x}_0$ is encoded into a hidden state $\bm{z}_0$. The control input $\bm{u}$ and context variable $\vbeta$ are encoded into feature vectors and passed as inputs to the GRU
\begin{align}
\bm{z}_0 &= e_{x}(\vx_{0}) \\
\mathbf{z}_{n+1} &= \operatorname{GRU} \left( \bm{z}_n, [e_{u}(\vu_n), e_{\beta}(\vbeta) ] \right) \label{eq:tradyn:gru}
\end{align}
where $e_{x}$, $e_{u}$, and $e_{\beta}$ are neural network encoders. 
The (predicted) latent state $\mathbf{z}_{n}$ is decoded into a Gaussian distribution in the state space
\begin{equation}
    \begin{split}
        \vx_{n} &\sim \mathcal{N} \left( d_{x,\mu}(\mathbf{z}_{n}), d_{x,\sigma^2}(\mathbf{z}_{n}) \right)
    \end{split}
\end{equation}
using neural networks $d_{x,\mu}$, $d_{x,\sigma^2}$.

The context encoder gets as input a set of state-action-state transitions $\sC^\alpha$ with flexible size $K$.
The context encoder is implemented by first encoding each transition in the context set independently using a transition encoder $e_{\mathrm{trans}}$, and, for permutation invariance, aggregating the encodings using a dimension-wise max operation. This yields the aggregated latent variable $\bm{z}_{\beta}$.
Lastly, a Gaussian density over the context variable~$\vbeta$ is predicted from the aggregated encodings
\begin{equation}
    \qctx( \vbeta \mid \sC^\alpha ) = \mathcal{N}\left( \vbeta; d_{\beta,\mu}( \vz_\beta ), \operatorname{diag}( d_{\beta,\sigma^2}( \vz_\beta ) ) \right)
\end{equation} with neural network decoders $d_{\beta,\mu}$, $d_{\beta,\sigma^2}$.
The network $d_{\beta,\sigma^2}$ is designed so that the predicted variance is positive and decreases monotonically when adding context observations.

To form a tractable loss, the marginal log likelihood in \cref{eq:tradyn:mll} is (approximately, see~\cite{le2018empirical}) bounded using the evidence lower bound
\begin{multline}
    \label{eq:tradyn:elbo}
    \log p( \lD^\alpha \mid \sC^\alpha ) \gtrapprox \mathbb{E}_{\vbeta \sim  \qctx( \vbeta \mid \lD^\alpha \cup \sC^\alpha )} \left[ \log p( \lD^\alpha \mid \vbeta ) \right]\\
    - \lambda_{\mathit{KL}} \operatorname{KL}\left( \qctx( \vbeta \mid  \lD^\alpha \cup \sC^\alpha )  \, \| \, \qctx( \vbeta \mid \sC^\alpha ) \right).
\end{multline}
similar to Neural Processes~\cite{garnelo2018neural}.
For training the dynamics model and context encoder, the approximate bound in \cref{eq:tradyn:elbo} is maximized by stochastic gradient ascent on empirical samples for target rollouts and context sets. Samples are drawn from trajectories generated on a training set of environments. 

By collecting context observations at test time, and inferring $\vbeta$ using $\qctx(\vbeta \mid \sC^\alpha)$, the dynamics model $\qfwd(\vx^{}_{n} \mid \vx^{}_{0}, \vu^{}_{0:N-1}, \vbeta)$ can adapt to a particular environment instance $\valpha$ (called \emph{calibration}).

  \begin{figure}
    \begin{tikzpicture}[
        scale=1,
        transform shape,
        line width=0.04cm,
        nx/.style={scale=1},
        nu/.style={scale=0.8},
        encoder/.style={
          draw, trapezium,
          trapezium stretches=true,
          trapezium left angle=60,
          trapezium right angle=60,
          scale=0.8,
          minimum height=0.7cm,
          draw=magenta!70!black,fill=magenta!20
        },
        enc_u/.style={
          draw, trapezium,
          trapezium stretches=true,
          trapezium left angle=60,
          trapezium right angle=60,
          scale=0.8,
          minimum height=0.35cm,
          draw=magenta!70!black,fill=magenta!20
        },
        decoder/.style={
          draw, trapezium,
          trapezium stretches=true,
          trapezium left angle=-60,
          trapezium right angle=-60,
          scale=0.8,
          minimum height=0.7cm,
          draw=green!70!black,fill=green!20
        },
        gru/.style={
          draw, rectangle,
          scale=0.8,
          minimum width=2cm,
          minimum height=0.7cm,
          draw=orange!70!black,fill=orange!20,rounded corners=2pt
        },
        ]
        \begin{scope}
          \node (x0) [nx] {$\bm{x}_0$};
          \node (enc_x) [encoder, below=0.4 of x0, anchor=north] {Encode};
          \node (circ_x) [circle, color=cyan!60, fill=cyan!20, very thick, below=0.7 of enc_x.south, anchor=center] {};
          \node (dec_x) [decoder, below=0.7 of circ_x.center, anchor=north] {Decode};
          \node (x0h) [nx, below=0.4 of dec_x.south, anchor=north] {\color{red}$\bm{\hat{x}}_0$};
          \node (gru_1) [gru, right=1 of circ_x.east, anchor=west] {GRU};
          \node (circ_x_2) [circle, color=red!60, fill=cyan!20, very thick, right=0.5 of gru_1.east, anchor=west] {};
          \node (gru_2) [gru, right=0.5 of circ_x_2.east, anchor=west] {GRU};
          
          \node (enc_t) [enc_u, above right=0.7 and 0.7 of gru_1.center, anchor=south,] {};
          \node (enc_u) [enc_u, above=0.7 of gru_1.center, anchor=south] {};
          \node (enc_b) [enc_u, above left=0.7 and 0.7 of gru_1.center, anchor=south] {};
          
          \node (b) [nu,above=0.2 of enc_b.north] {$\bm{\beta}$};
          \node (u) [nu,above=0.2 of enc_u.north] {$\bm{u}_0$};
          \node (t) [nu,above=0.2 of enc_t.north] {${\tau}(\bm{x}_0)$};
          
          \node (enc_t_2) [enc_u, above right=0.7 and 0.7 of gru_2.center, anchor=south,] {};
          \node (enc_u_2) [enc_u, above=0.7 of gru_2.center, anchor=south] {};
          \node (enc_b_2) [enc_u, above left=0.7 and 0.7 of gru_2.center, anchor=south] {};
          
          \node (b2) [nu,above=0.2 of enc_b_2.north] {$\bm{\beta}$};
          \node (u2) [nu,above=0.2 of enc_u_2.north] {$\bm{u}_1$};
          \node (t2) [nu,above=0.2 of enc_t_2.north] {${\tau}(\bm{x}_1)$};
          \node (t2h) [nu,above=0 of t2] {${\tau}({\color{red}\bm{\hat{x}}_1})$};
          \draw[dotted] ($0.5*(t2.north west)+0.5*(t2h.south west)$) -- ([xshift=2.3cm]$0.5*(t2.north west)+0.5*(t2h.south west)$);
          
          \node (t2tt) [nu,right=0.07 of t2.east, anchor=west] {\emph{training}};
          \node (t2tp) [nu,right=0.07 of t2h.east, anchor=west] {\emph{prediction}};
          
          \node (circ_x_3) [circle, color=red!60, fill=cyan!20, very thick, right=0.5 of gru_2.east, anchor=west] {};
          
          \node (circ_x_4) [circle, color=red!60, fill=cyan!0, very thick, right=0.5 of circ_x_3.east, anchor=west] {};
          
          \node (dec_x_2) [decoder, below=0.7 of circ_x_2.center, anchor=north] {Decode};
          
          \node (dec_x_3) [decoder, below=0.7 of circ_x_3.center, anchor=north] {Decode};
          
          \node (x1h) [nx, below=0.4 of dec_x_2.south, anchor=north] {\color{red}$\bm{\hat{x}}_1$};
          
          \node (x2h) [nx, below=0.4 of dec_x_3.south, anchor=north] {\color{red}$\bm{\hat{x}}_2$};
          
          \draw[-stealth] (x0) -- (enc_x.north);
          \draw[-stealth] (enc_x.south) -- (circ_x.north);
          \draw[-stealth] (circ_x.south) -- (dec_x.north);
          \draw[-stealth] (dec_x.south) -- (x0h.north);
          
          \draw[-stealth] (circ_x.east) -- (gru_1.west);
          
          \draw[-stealth] (t.south) -- (enc_t.north);
          \draw[-stealth] (u.south) -- (enc_u.north);
          \draw[-stealth] (b.south) -- (enc_b.north);
          
          \draw[-stealth] (enc_t.south) -- (gru_1.north);
          \draw[-stealth] (enc_u.south) -- (gru_1.north);
          \draw[-stealth] (enc_b.south) -- (gru_1.north);
          
          \draw[-stealth] (t2.south) -- (enc_t_2.north);
          \draw[-stealth] (u2.south) -- (enc_u_2.north);
          \draw[-stealth] (b2.south) -- (enc_b_2.north);
          
          \draw[-stealth] (enc_t_2.south) -- (gru_2.north);
          \draw[-stealth] (enc_u_2.south) -- (gru_2.north);
          \draw[-stealth] (enc_b_2.south) -- (gru_2.north);
          
          \draw[-stealth] (gru_1.east) -- (circ_x_2.west);
          \draw[-stealth] (circ_x_2.east) -- (gru_2.west);
          
          \draw[-stealth] (circ_x_2.south) -- (dec_x_2.north);
          \draw[-stealth] (circ_x_3.south) -- (dec_x_3.north);
          
          \draw[-stealth] (dec_x_2.south) -- (x1h.north);
          \draw[-stealth] (dec_x_3.south) -- (x2h.north);
          
          \draw[-stealth] (gru_2.east) -- (circ_x_3.west);
          
          \draw[-stealth, dotted] (circ_x_3.east) -- (circ_x_4.west);
        \end{scope}
      \end{tikzpicture}
      \caption{Architecture of our proposed terrain- and robot-aware forward dynamics model (TRADYN). The initial state of the robot $\bm{x}_0$ is embedded as hidden state of a gated recurrent unit (GRU) cell. The GRU makes a single-step forward prediction in the latent space using embeddings of the context variable $\bm{\beta}$, action $\bm{u}$ and terrain observation $\bm{\tau}$ as additional inputs. Latent states are mapped to Gaussian distributions on the robot's observation space for decoding. While during training the actual terrain observation ${\tau}(\bm{x}_n)$ is used, during prediction, the map ${\tau}$ is queried at predicted robot locations ${\tau}({\color{red}\bm{\hat{x}}_n})$. See \cref{sec:tradyn:method} for details.}
      \label{fig:tradyn:model}
  \end{figure}
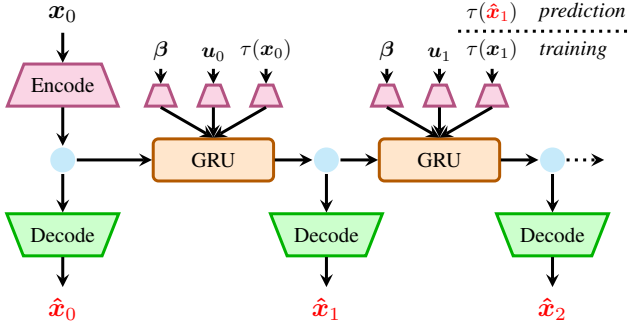

\section{METHOD}
\label{sec:tradyn:method}
In the modeling assumption of EtC, changes in the dynamics among different instances of environments are captured in a global latent variable $\valpha$ (see \cref{eq:tradyn:etc_assumption}) which is unobserved.
In terrain-aware robot navigation, among different environments, the terrain varies (with the terrain layout captured by $\aterrfcn$), in addition to robot-specific parameters such as actuator gains (captured by $\arobot$).
In principle, both effects can be absorbed into a single latent variable $\valpha = (\aterrfcn, \arobot)$.
Here, we make more specific assumptions, and assume the terrain-specific properties to be captured in a state-dependent function $\aterrfcn(\vx_n)$.


\subsection{Terrain- and Robot-Aware Dynamics Model}
Conclusively, we assume the following environment dynamics
\begin{equation}
    \vx_{n+1} = f( \vx_n, \vu_n, \arobot, \aterrfcn(\vx_n)) + \veps_n
\end{equation}
with $\veps_n \sim \mathcal{N}( 0, \mQ_n)$ as in \cref{eq:tradyn:etc_assumption}.
In our case of terrain-aware robot navigation, $\vx_n$ refers to the robot state at timestep~$n$, $\vu_n$ are the control inputs, $\arobot$ captures (unobserved) properties of the robot (mass, actuator gains), and $\aterrfcn(\vx_n)$ captures the spatially dependent terrain properties (e.g., friction). While we assume $\aterrfcn$ to be unobserved, we assume the existence of a \emph{known map} of \emph{terrain features} $\tterrain(\vx_n)$, which can be queried at any $\bm{x}_n$ to estimate the value of $\aterrfcn(\vx_n)$.
Exemplarily, $\tterrain$ may yield visual terrain observations, which relate to friction coefficients. 

As we retain the assumption of EtC that $\arobot$ is not directly observable, we condition the multi-step forward dynamics model on the latent variable $\bm{\beta}$. In addition, we condition on observed terrain features $\bm{\tau}_{0:n-1}$, i.e.,
\begin{equation}
    \bm{\hat{x}}_{n} \sim \qfwd(\bm{x}_{n} \mid \bm{x}_{0}, \bm{u}_{0:n-1}, \bm{\beta}, \bm{\tau}_{0:n-1}).
\end{equation}
We obtain $\bm{\tau}_{0:n-1}$ differently for training and prediction. During training, we evaluate ${\tau}$ at ground-truth states, i.e. $\bm{\tau}_{0} = {\tau}(\bm{x}_{0}), \bm{\tau}_{1} = {\tau}(\bm{x}_{1})$, etc. During prediction, we do not have access to ground-truth states, and obtain $\bm{\tau}_{0:n-1}$ auto-regressively from predictions as $\bm{\tau}_{0} = {\tau}(\bm{x}_{0}), \bm{\tau}_{1} = {\tau}(\bm{\hat{x}}_{1})$, etc. 

To capture terrain-specific properties, we extend EtC as follows. We introduce an additional encoder $e_\tau$ which encodes a terrain feature $\bm{\tau}$. The encoded value is passed as input to the GRU, such that \cref{eq:tradyn:gru} is updated to 
\begin{equation}
    \mathbf{z}_{n+1} = \operatorname{GRU} \left( \bm{z}_0, [e_{\tau}(\boldsymbol{\tau}_n), e_{u}(\vu_n), e_{\beta}(\vbeta) ] \right).
\end{equation}
Also, the context set is extended to contain terrain features
\begin{equation}
\begin{aligned}
\sC^\alpha = {\{(\vx^{(k)}, {\tau}(\vx^{(k)}),  \vu^{(k)}, \vx_{+}^{(k)}, {\tau}(\vx_{+}^{(k)}))\}}_{k=1}^K.
\end{aligned}
\end{equation}
We refer to \cref{fig:tradyn:model} for a depiction of our model.

For each training example, the context set size $K$ is uniformly sampled in $\{0,\ldots,50\}$. The target rollout length is $N=50$. As in EtC~\cite{achterhold2021explore}, we set $\lambda_{\mathit{KL}} = 5$. The dimensionality of the latent variable $\vbeta$ is $16$. For details on the networks $(e_u, e_\beta, d_{x,\mu}, d_{x,\sigma^2}, e_\mathrm{trans}, d_{\beta,\mu}, d_{\beta,\sigma^2})$ we refer to~\cite{achterhold2021explore}, as we strictly follow the architecture described therein. 
The additional encoder network we introduce, $e_{\beta}$, follows the architecture of $e_{\tau}$ and $e_{u}$. It contains a single hidden layer with 200 units and ReLU activations, and an output layer which maps to an embedding of dimensionality $200$.

\subsection{Path Planning and Motion Control}
\label{sec:tradyn:pathplanning}

We use TRADYN in a model-predictive control setup. The model $\qfwd$ yields state predictions $\bm{\hat{x}}_{1:H}$ for an initial state $\bm{x_0}$ and controls $\bm{u}_{0:H-1}$. For calibration, i.e., inferring $\bm{\beta}$ from a context set $\sC$ with the context encoder $\qctx$, calibration transitions are collected on the target environment prior to planning. This allows adapting to varying robot parameters. The predictive terrain feature lookup (see \cref{fig:tradyn:model}) with ${\tau}(\vx)$ allows adapting to varying terrains. 
We use the Cross-Entropy Method (CEM~\cite{rubinstein1999cross}) for planning.
We aim to reach the target position with minimal throttle control energy, given by the sum of squared throttle commands during navigation. This gives rise to the following planning objective, which penalizes high throttle control energy and a deviation of the robot's terminal position to the target position $\bm{p}^*$:
\begin{multline}
    \label{eq:tradyn:cost}
     J(\bm{u}_{0:H-1}, \bm{\hat{x}}_{1:H}) = \\ \frac{1}{2} \sum_{n=0}^{H-1} u_{\mathrm{throttle}, n}^2 + || {[\hat{p}_{\mathrm{x},H}, \hat{p}_{\mathrm{y},H}]}^\top - \bm{p}^* ||_2^2.
\end{multline}
In our CEM implementation, we normalize the distance term in \cref{eq:tradyn:cost} to have zero mean and unit variance over all CEM candidates, to trade-off control- and distance cost terms even under large terrain variations.
At each step, we only apply the first action and plan again from the resulting state in a receding horizon scheme.

\section{EXPERIMENTS}

\begin{figure}[tb]
    \centering
    \includegraphics[scale=1]{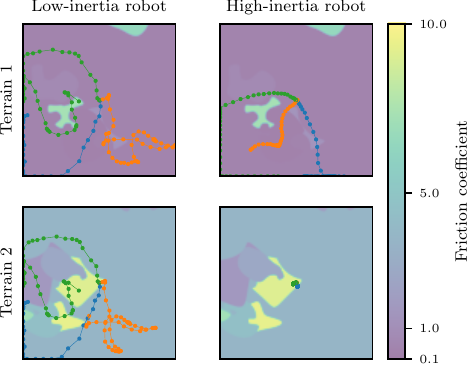}
    \caption{Exemplary rollouts (length 50) on two different terrain layouts (rows) and for two exemplary robot configurations (low-inertia, high-inertia) (columns). Rollouts start from the center; actions are sampled time-correlated. The low-inertia robot has minimal mass $m=1$ and maximal control gains $k_\mathrm{throttle}=1000$, $k_\mathrm{steer}=\pi/4$. The high-inertia robot has maximal mass $m=4$ and minimal control gains $k_\mathrm{throttle}=500$, $k_\mathrm{steer}=\pi/8$. Equally colored trajectories ({\scriptsize \color{mpl_propcycle_1} \newmoon}, {\scriptsize \color{mpl_propcycle_2} \newmoon}, {\scriptsize \color{mpl_propcycle_3} \newmoon}) correspond to identical sequences of applied actions. See \cref{sec:tradyn:simenv} for details.}
    \label{fig:tradyn:rollouts}
\end{figure}

\begin{figure}[tb]
    \centering
    \includegraphics[scale=1]{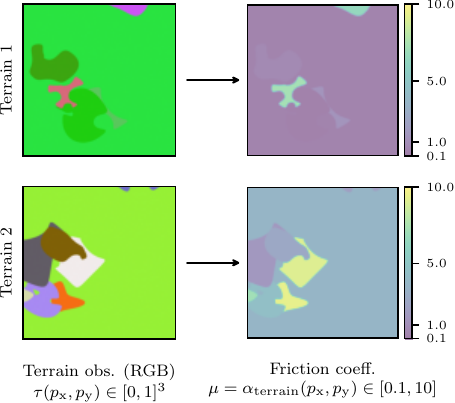}
    \caption{Relationship of RGB terrain features $\bm{\tau}$ (left column) to friction coefficient $\mu$ (right column). See \cref{sec:tradyn:terrainlayouts} for details.}
    \label{fig:tradyn:terrains}
\end{figure}

\begin{figure*}[tb!]
    \centering
    \includegraphics[width=\linewidth]{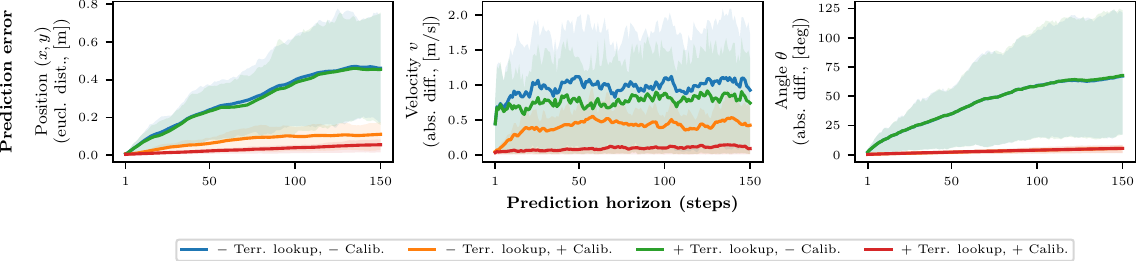}
    \caption{Prediction error evaluation for the proposed model and its ablations (no terrain lookup / no calibration), plotted over the prediction horizon (number of prediction steps). From left to right: Positional error (euclidean distance), velocity error (absolute difference), angular error (absolute difference). Depicted are the mean and 20\%, 80\% percentiles over 150 evaluation rollouts for 5 independently trained models per model variant. Our approach with terrain lookup and calibration clearly outperforms the other variants in position and velocity prediction (left and center panel). For predicting the angle (right panel), terrain friction is not relevant, which is why the terrain lookup brings no advantage. However, calibration is important for accurate angle prediction. See \cref{sec:tradyn:eval_prediction} for details.}
    \label{fig:tradyn:unicycle_prediction}
\end{figure*}

\begin{figure}[tb!]
    \centering
    \includegraphics[scale=1]{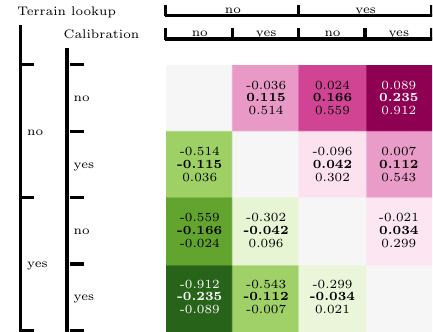}
    \caption{Comparison of model variants with and without terrain lookup and calibration. $E^v_{k,i}$ denotes the throttle control energy for method $v$ on navigation task $k \in \{1, \ldots, 150\}$ for a trained model with seed $i \in \{1, \ldots, 5\}$. We show statistics (20\% percentile, median, 80\% percentile) on the set of pairwise comparisons of control energies $\{ 
    E^{\mathrm{row}}_{k,i_1} - E^{\mathrm{col}}_{k,i_2} \mid \forall k \in \{1,\ldots,K\}, i_1 \in \{1,\ldots,5\}, i_2 \in \{1,\ldots,5\}
    \}$. Significant ($p<0.05$) results are printed \textbf{bold} (see \cref{sec:tradyn:eval_planning}). Exemplarily, both performing terrain lookup and calibration (last row) yields navigation solutions with significantly lower throttle control energy (negative numbers) compared to all other methods (columns). See \cref{sec:tradyn:eval_planning} for details.
    }
    \label{fig:tradyn:unicycle_planning_gain}
\end{figure}

\begin{figure}[tb!]
    \centering
    \includegraphics[scale=1]{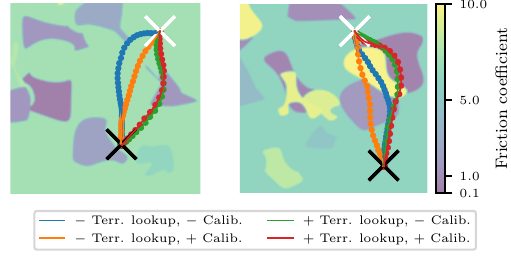}\\
    \vspace{1em}
    {\footnotesize
    \begin{tabular}{lcccc} 
    \toprule
    & \multicolumn{2}{c}{left terrain} & \multicolumn{2}{c}{right terrain} \\
     \cmidrule(lr){2-3} \cmidrule(lr){4-5}
    Variant & \makecell{Thr.\ ctrl.\\energy} & \makecell{Target\\dist [\SI{}{mm}]} & \makecell{Thr.\ ctrl.\\energy} & \makecell{Target\\dist [\SI{}{mm}]}\\
    \midrule 
{\scriptsize \color{mpl_propcycle_1} \newmoon} $-$T, $-$C & 4.08 & 7.82 & 3.96 & 2.69 \\
        {\scriptsize \color{mpl_propcycle_2} \newmoon} $-$T, $+$C& 3.47 & 4.89 & 2.78 & 4.87 \\
        {\scriptsize \color{mpl_propcycle_3} \newmoon} $+$T, $-$C & 2.01 & 4.57 & 1.88 & 5.14\\
        {\scriptsize \color{mpl_propcycle_4} \newmoon} $+$T, $+$C & 2.02 & 5.55 & 1.43 & 6.66 \\
    \bottomrule
    \end{tabular}
    }
    \caption{Exemplary navigation trajectories and their associated throttle control energy and final distance to the target (see table). The robot starts at the white cross, the goal is marked by a black cross. With terrain lookup ({\scriptsize \color{mpl_propcycle_3} \newmoon} +T, -C and  {\scriptsize \color{mpl_propcycle_4} \newmoon} +T, +C), our method circumvents areas of high friction coefficient (i.e., high energy dissipation), resulting in lower throttle control energy (see table). Enabling calibration (+C) further reduces throttle control energy on the right terrain. See \cref{sec:tradyn:eval_planning} for details.}
    \label{fig:tradyn:unicycle_planning_examples}
\end{figure}

\begin{figure}[tb!]
    \centering
    \includegraphics[width=0.99\linewidth]{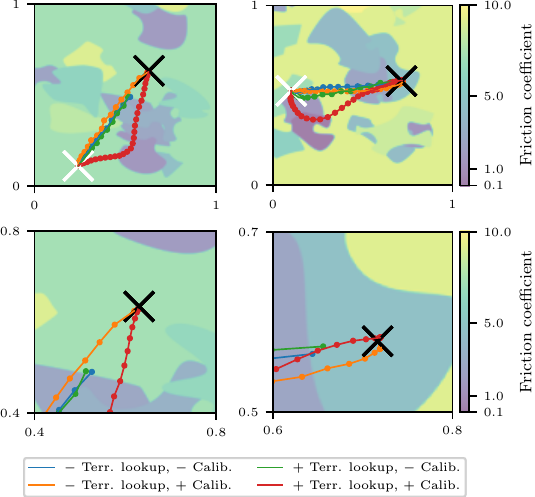}\\
    \caption{Failure cases for the non-calibrated models. The top row shows the full terrain of extent $[0, 1]$\SI{}{\meter}. The bottom row is zoomed around the goal. In the cases shown, planning with the non-calibrated models does not succeed in reaching the goal marked by the black cross within the given step limit of $50$ steps, in contrast to the calibrated models. See \cref{sec:tradyn:eval_planning} for details.}
    \label{fig:tradyn:unicycle_planning_fail}
\end{figure}


\subsection{Simulation environment}
\label{sec:tradyn:simenv}

\subsubsection{Simulated Robot Dynamics}
\label{sec:tradyn:simenv_robdyn}

We perform experiments in a 2D simulation with a unicycle-like robot setup where the continuous time-variant 2D dynamics with position $\mathbf{p} = {[\cpx, \cpy]}^\top$, orientation $\varphi'$, and directional velocity $v$, for control input $\bm{u} = {[u_\mathrm{throttle}, u_\mathrm{steer}]}^\top \in {[-1, 1]}^2$, are given by

\begin{equation}
    \label{eq:tradyn:continuous_time_turtlebot_dynamics}
    \begin{aligned}
        \dot{\bm{p}}(t) & = {\begin{bmatrix} \cos\varphi' & \sin\varphi' \end{bmatrix}}^{\top} {v}(t) \\
        \dot{{v}}(t) & = \frac{1}{m} (F_\mathrm{throttle} + F_\mathrm{fric}) \\
        F_\mathrm{throttle} & = u_\mathrm{throttle} \: k_\mathrm{throttle} \\
        F_\mathrm{fric} & = -\operatorname{sign}(v(t)) \: \mu \: m \: g. \\
    \end{aligned}
\end{equation}

As our method does not use continuous-time observations, but only discrete-time samplings with stepsize $\Delta_T = \SI{0.01}{\second}$, we approximate the state evolution between two timesteps as follows. First, we apply the change in angle as $\varphi' = \varphi(t + \Delta_T) = \varphi(t) + u_\mathrm{steer} \: k_\mathrm{steer}$. 
We then query the terrain friction coefficient $\mu$  at the position $\bm{p}(t)$.
With the friction coefficient $\mu$ and angle $\varphi'$ we compute the evolution of position and velocity with \cref{eq:tradyn:continuous_time_turtlebot_dynamics}.
The existence of the friction term in \cref{eq:tradyn:continuous_time_turtlebot_dynamics} requires an accurate integration, which is why we solve the initial value problem in \cref{eq:tradyn:continuous_time_turtlebot_dynamics} numerically using an explicit Runge Kutta (RK45) method, yielding ${\bm{p}}(t+\Delta_T)$ and ${{v}}(t+\Delta_T)$.
Our simulated system dynamics are deterministic.
To avoid discontinuities, we represent observations of the above system as $\bm{x}(t) = {[\cpx(t), \cpy(t), v(t)/5, \cos \varphi(t), \sin \varphi(t)]}^\top$. We use $g=\SI{9.81}{\metre\per\square\second}$ as gravitational acceleration. Positions $\bm{p}(t)$ are clipped to the range $[0, 1] \: \SI{}{\metre}$; the directional velocity $v(t)$ is clipped to $[-5, 5] \: \SI{}{\metre\per\second}$. The friction coefficient $\mu = \aterrfcn(\cpx, \cpy) \in [0.1, 10]$ depends on the terrain layout $\aterrfcn$ and the robot's position. 
The mass $m$ and control gains $k_\mathrm{throttle}, k_\mathrm{steer}$ are robot-specific properties, we refer to \cref{table:tradyn:robot_specific_properties_setup} for their value ranges.

\subsubsection{Terrain layouts}
\label{sec:tradyn:terrainlayouts}
To simulate the influence of varying terrain properties on the robots' dynamics, we programmatically generate 50 terrain layouts for training the dynamics model and 50 terrain layouts for testing (i.e., in prediction- and planning evaluation). For generating terrain $k$, we first generate an unnormalized feature map $\hat{\tau}^{(k)}$, from which we compute $\aterrfcn^{(k)}$ and the normalized feature map ${\tau}^{(k)}$.
The unnormalized feature map is represented by a 2D RGB image of size $460\,\mathrm{px} \times 460\,\mathrm{px}$.
For its generation, first, a background color is randomly sampled, followed by sequentially placing randomly sampled patches with cubic bezier contours.
The color value $(r, g, b) \in {\{0, \ldots, 255\}}^3$ at each pixel maps to the friction coefficient $\mu = \alpha_\mathrm{terrain}(\cpx,\cpy)$ through bitwise left-shifts $\ll$ as
\begin{equation}
    \label{eq:tradyn:tau_to_mu}
    \begin{split}
    & \eta =  {((r \ll 16) + (g \ll 8) + b)}/{(2^{24}-1)} \\
    & \alpha_\mathrm{terrain}(\cpx,\cpy) = 0.1 + (10-0.1)\eta^2.
    \end{split}
\end{equation}
The agent can observe the normalized terrain color $\tau^{(k)}(\cpx, \cpy) \in {[0, 1]}^3$ with $\tau^{(k)}(\cpx, \cpy) = \hat{\tau}^{(k)}(\cpx, \cpy) / 255$, and can query $\tau^{(k)}(\cpx, \cpy)$ at arbitrary $\cpx, \cpy$.
The simulator has direct access to $\mu = \aterrfcn^{(k)}(\cpx, \cpy)$.
We denote the training set of terrains $\mathcal{A}_\mathrm{train} = \{ \aterrfcn^{(k)} \mid k \in \{1,\ldots,50\} \}$ and the test set of terrains $\mathcal{A}_\mathrm{test} = \{ \aterrfcn^{(k)} \mid k \in \{51,\ldots,100\} \}$.
We refer to \cref{fig:tradyn:terrains} for a visualization of two terrains and the related friction coefficients.

\subsubsection{Environment instance}
The robot's dynamics depends on the terrain $\aterrfcn$ as it is the position-dependent friction coefficient, and the robot-specific parameters $\arobot = (m, k_\mathrm{throttle}, k_\mathrm{steer})$. A fixed tuple $(\aterrfcn,\arobot)$ forms an environment \emph{instance}.

\subsubsection{Trajectory generation}
We require the generation of trajectories at multiple places of our algorithm for training and evaluation: To generate training data, to sample candidate trajectories for the cross-entropy planning method, to generate calibration trajectories, and to generate trajectories for evaluating the prediction performance. One option would be to generate trajectories by independently sampling actions from a Gaussian distribution at each timestep. However, this Brownian random walk significantly limits the space traversed by such trajectories~\cite{pinneri2020sample}. To increase the traversed space,~\cite{pinneri2020sample} propose to use time-correlated (colored) noise with a power spectral density $\operatorname{PSD}(f) \propto \frac{1}{f^\omega}$, where $f$ is the frequency. We use $\omega=0.5$ in all our experiments.

\subsubsection{Exemplary rollouts} 
\label{par:tradyn:exemplaryrollouts}
We visualize exemplary rollouts on different terrains and with different robot parametrizations in \cref{fig:tradyn:rollouts}. We observe that both the terrain-dependent friction coefficient $\mu$, as well as the robot properties, have a significant influence on the shape of the trajectories, highlighting the importance of a model to be able to adapt to these properties.

\begin{table}[t]
\begin{center}
\caption{Robot-specific properties.}
\label{table:tradyn:robot_specific_properties_setup}
\begin{tabular}{lcc}
\toprule
\multicolumn{1}{l}{ Property} &\multicolumn{1}{c}{ Min.} &\multicolumn{1}{c}{Max.} \\
\midrule
Mass $m$ [kg]          & 1    & 4\\
Throttle gain $k_\mathrm{throttle}$          &500  &1000\\
Steer gain $k_\mathrm{steer}$      & $\pi$/8   & $\pi$/4\\
\bottomrule
\end{tabular}
\end{center}
\vspace{-1em}
\end{table}

\subsection{Model training}
\label{sec:tradyn:modeltraining}

We train our proposed model on a set of precollected trajectories on different terrain layouts and robot parametrizations. 
First, we sample a set of $10000$ unique terrain layout / robot parameter settings to generate training trajectories. For validation, a set of additional 5000 settings is used. On each setting, we generate two trajectories, used later during training to form the target rollout $\lD^\alpha$ and context set $\sC^\alpha$, respectively.
Terrain layouts are sampled uniformly from the training set of terrains, i.e. $\mathcal{A}_\mathrm{train}$. Robot parameters are sampled uniformly from the parameter ranges given in \cref{table:tradyn:robot_specific_properties_setup}.
The robots' initial state $\bm{x_0} = {[p_{\mathrm{x},0}, p_{\mathrm{y},0}, v_0, \varphi_0]}^\top$ is uniformly sampled from the ranges $p_{\mathrm{x},0}, p_{\mathrm{y},0} \in [0, 1]$, $v_0 \in [-5, 5]$, $\varphi_0 \in [0, 2\pi]$. Each trajectory consists of $100$ applied actions and the resulting states. We use time-correlated (colored) noise to sample actions (see previous paragraph). We follow the training procedure by~\cite{achterhold2021explore}, except we evaluate the models after 100k training steps.

\subsubsection{Model ablation}
As an ablation to our model, we only input the terrain features $\bm{\tau}$ at the current and previously visited states of the robot as terrain observations, but do not allow for terrain lookups  in a map at future states  during prediction. We will refer to this ablation as \emph{No($-$) terrain lookup} in the following.

\subsection{Prediction evaluation}
\label{sec:tradyn:eval_prediction}
In this section we evaluate the prediction performance of our proposed model. To this end, we generate 150 test trajectories of length 150, on the \emph{test} set of terrain layouts $\mathcal{A}_\mathrm{test}$. Robot parameters are uniformly sampled as during data collection for model training. The robot's initial position is sampled from ${[0.1, 0.9]}^2$, the orientation from $[0, 2\pi]$. The initial velocity is fixed to 0. Actions are sampled with a time-correlated (colored) noise scheme. In case the model is \emph{calibrated}, we additionally collect a small trajectory for each trajectory to be predicted, consisting of 10 transitions, starting from the same initial state $\bm{x_0}$, but with different random actions. Transitions from this trajectory form the context set $\sC$, which is used by the context encoder $\qctx(\bm{\beta} \mid \sC)$ to output a belief on the latent context variable~$\bm{\beta}$. In case the model is \emph{not calibrated}, the distribution is given by the context encoder for an empty context set, i.e. $\qctx(\bm{\beta} \mid \sC = \{\})$. We evaluate two model variants; first, our proposed model which utilizes the terrain map~${\tau}(\cpx, \cpy)$ for lookup during predictions, and second, a model for which the terrain observation is concatenated to the robot observation.
All results are reported on 5 independently trained models.
\Cref{fig:tradyn:unicycle_prediction} shows that our approach with terrain lookup and calibration clearly outperforms the other variants in position and velocity prediction. As the evolution of the robot's angle is independent of terrain friction, for angle prediction, only performing calibration is important.


\subsection{Planning evaluation}
\label{sec:tradyn:eval_planning}
Aside the prediction capabilities of our proposed method, we are interested whether it can be leveraged for efficient navigation planning. To evaluate the planning performance, we generate 150 navigation tasks, similar to the above prediction tasks, but with an additional randomly sampled target position $\bm{p}^* \in {[0.1,0.9]}^2$ for the robot. We perform receding horizon control as described in \cref{sec:tradyn:pathplanning}.

Again, we evaluate four variants of our model. We compare models with and without the ability to perform terrain lookups. Additionally, we evaluate the influence of calibration, by either collecting 10 additional calibration transitions for each planning task setup, or not collecting any calibration transition ($\sC = \{ \}$), giving four variants in total.

As we have trained five models with different seeds, over all models, we obtain 750 navigation results. We count a navigation task as \emph{failed} if the final Euclidean distance to the goal exceeds $\SI{5}{\centi\meter}$. 

We evaluate the efficiency of the navigation task solution by the sum of squared throttle controls over a fixed trajectory length of $N=50$ steps, which we denote as $E = \sum_{n=0}^{N-1} u_{\mathrm{throttle}, n}^2$.
We introduce super- and subscripts $E^v_{k,i}$ to refer to model variant $v$, planning task index $k$ and model seed $i$. Please see \cref{fig:tradyn:unicycle_planning_gain,fig:tradyn:unicycle_planning_examples} for results comparing the particular variants.  
For pairwise comparison of control energies $E$ we leverage the Wilcoxon signed-rank test with a $p$-value of $0.05$.
We can conclude that, regardless of calibration, performing terrain lookups yields navigation solutions with significantly lower throttle control energy. The same holds for performing calibration, regardless of performing terrain lookups. Lowest control energy is obtained for both performing calibration and terrain lookup.

We refer to \cref{table:tradyn:planning_statistics} for statistics on the number of failed tasks and final distance to the goal. 
As can be seen, our terrain- and robot-aware approach yields overall best performance in Euclidean distance to the goal and succeeds in all runs in reaching the goal. Planning with non-calibrated models variants occasionally fails, i.e., the goal is not reached. We show such failure cases in \cref{fig:tradyn:unicycle_planning_fail}.

\begin{table}[tb]
\centering
\caption{Distance to goal (median and 20\% / 80\% percentiles) and failure rate for \SI{5}{\centi\meter} distance threshold to goal. Variants are with/without terrain lookup ($\pm$T) and with/without calibration ($\pm$C). Our full approach ($+$T, $+$C) yields best performance in reaching the goal and succeeds in all runs.}
\label{table:tradyn:planning_statistics}
\begin{tabular}{lcccc} 
    \toprule
    Variant & \multicolumn{3}{c}{Euclidean distance to goal [\SI{}{\milli\meter}]} & Failed tasks\\
    & P20 & median & P80 & \\
    \midrule 
{\scriptsize \color{mpl_propcycle_1} \newmoon} $-$T, $-$C & 3.00 & 5.19 & 8.67 &  6/750 \\
        {\scriptsize \color{mpl_propcycle_2} \newmoon} $-$T, $+$C & 3.13 & 5.23 & 7.65 &\textbf{0/750} \\
        {\scriptsize \color{mpl_propcycle_3} \newmoon} $+$T, $-$C & 2.33 & 4.22 & 6.49 &14/750 \\
        {\scriptsize \color{mpl_propcycle_4} \newmoon} $+$T, $+$C & \textbf{2.16} & \textbf{3.85} & \textbf{5.61} & \textbf{0/750} \\
    \bottomrule
\end{tabular}
\end{table}

\section{CONCLUSIONS}
In this paper, we propose a forward dynamics model which can adapt to variations in unobserved variables that govern the system's dynamics such as robot-specific properties, as well as to {spatial} variations. We train our model on a simulated unicycle-like robot, which has varying mass and actuator gains. In addition, the robot's dynamics are influenced by instance-wise and spatially varying friction coefficients of the terrain, which are only indirectly observable through terrain observations. 
In 2D simulation experiments, we demonstrate that our model can successfully cope with such variations through calibration and terrain lookup. It exhibits smaller prediction errors compared to model variants without calibration and terrain lookup, and yields solutions to navigation tasks which require lower throttle control energy. 
In future work, we plan to extend our novel learning-based approach for real-world robot navigation problems with partial observability, noisy state transitions, and noisy observations.





\bibliography{IEEEexample}

\begin{thebibliography}{10}
\providecommand{\url}[1]{#1}
\csname url@rmstyle\endcsname
\providecommand{\newblock}{\relax}
\providecommand{\bibinfo}[2]{#2}
\providecommand\BIBentrySTDinterwordspacing{\spaceskip=0pt\relax}
\providecommand\BIBentryALTinterwordstretchfactor{4}
\providecommand\BIBentryALTinterwordspacing{\spaceskip=\fontdimen2\font plus
\BIBentryALTinterwordstretchfactor\fontdimen3\font minus
  \fontdimen4\font\relax}
\providecommand\BIBforeignlanguage[2]{{%
\expandafter\ifx\csname l@#1\endcsname\relax
\typeout{** WARNING: IEEEtran.bst: No hyphenation pattern has been}%
\typeout{** loaded for the language `#1'. Using the pattern for}%
\typeout{** the default language instead.}%
\else
\language=\csname l@#1\endcsname
\fi
#2}}

\bibitem{sonker2021adding}
R.~Sonker and A.~Dutta, ``Adding terrain height to improve model learning for
  path tracking on uneven terrain by a four wheel robot,'' \emph{{IEEE}
  Robotics Autom. Lett.}, 2021.

\bibitem{valada2017adapnet}
A.~Valada, J.~Vertens, A.~Dhall, and W.~Burgard, ``Adapnet: Adaptive semantic
  segmentation in adverse environmental conditions,'' in \emph{{IEEE} {ICRA}},
  2017.

\bibitem{yang1028unifying}
K.~Yang, L.~M. Bergasa, E.~Romera, R.~Cheng, T.~Chen, and K.~Wang, ``Unifying
  terrain awareness through real-time semantic segmentation,'' in \emph{{IEEE}
  Intelligent Vehicles Symposium}, 2018.

\bibitem{zhu2019offroad}
Z.~Zhu, N.~Li, R.~Sun, H.~Zhao, and D.~Xu, ``Off-road autonomous vehicles
  traversability analysis and trajectory planning based on deep inverse
  reinforcement learning,'' \emph{CoRR}, vol. abs/1909.06953, 2019.

\bibitem{kahn2021badgr}
G.~Kahn, P.~Abbeel, and S.~Levine, ``{BADGR:} an autonomous self-supervised
  learning-based navigation system,'' \emph{{IEEE} Robotics Autom. Lett.},
  2021.

\bibitem{grigorescu2021lvdnmpc}
S.~Grigorescu, C.~Ginerica, M.~Zaha, G.~Macesanu, and B.~Trasnea, ``{LVD-NMPC}:
  A learning-based vision dynamics approach to nonlinear model predictive
  control for autonomous vehicles,'' \emph{International Journal of Advanced
  Robotic Systems}, 2021.

\bibitem{siva2021enhancing}
S.~Siva, M.~B. Wigness, J.~G. Rogers, and H.~Zhang, ``Enhancing consistent
  ground maneuverability by robot adaptation to complex off-road terrains,'' in
  \emph{CoRL}, 2021.

\bibitem{xiao2021learning}
X.~Xiao, J.~Biswas, and P.~Stone, ``Learning inverse kinodynamics for accurate
  high-speed off-road navigation on unstructured terrain,'' \emph{{IEEE}
  Robotics Autom. Lett.}, 2021.

\bibitem{sikand2022visual}
K.~S. Sikand, S.~Rabiee, A.~Uccello, X.~Xiao, G.~Warnell, and J.~Biswas,
  ``Visual representation learning for preference-aware path planning,'' in
  \emph{IEEE {ICRA}}, 2022.

\bibitem{deisenroth2011pilco}
M.~P. Deisenroth and C.~E. Rasmussen, ``{PILCO:} {A} model-based and
  data-efficient approach to policy search,'' in \emph{{ICML}}, 2011.

\bibitem{lenz2015deepmpc}
I.~Lenz, R.~A. Knepper, and A.~Saxena, ``Deep{MPC}: Learning deep latent
  features for model predictive control,'' in \emph{Robotics: Science and
  Systems}, 2015.

\bibitem{oh2015action}
J.~Oh, X.~Guo, H.~Lee, R.~L. Lewis, and S.~Singh, ``Action-conditional video
  prediction using deep networks in atari games,'' in \emph{NeurIPS}, 2015.

\bibitem{finn2016unsupervised}
C.~Finn, I.~J. Goodfellow, and S.~Levine, ``Unsupervised learning for physical
  interaction through video prediction,'' in \emph{NeurIPS}, 2016.

\bibitem{hafner2019learning}
D.~Hafner, T.~P. Lillicrap, I.~Fischer, R.~Villegas, D.~Ha, H.~Lee, and
  J.~Davidson, ``Learning latent dynamics for planning from pixels,'' in
  \emph{{ICML}}, 2019.

\bibitem{nagabandi2018neural}
A.~Nagabandi, G.~Kahn, R.~S. Fearing, and S.~Levine, ``Neural network dynamics
  for model-based deep reinforcement learning with model-free fine-tuning,'' in
  \emph{IEEE {ICRA}}, 2018.

\bibitem{shaj2020acrnn}
V.~Shaj, P.~Becker, D.~B{\"{u}}chler, H.~Pandya, N.~van Duijkeren, C.~J.
  Taylor, M.~Hanheide, and G.~Neumann, ``Action-conditional recurrent kalman
  networks for forward and inverse dynamics learning,'' in \emph{CoRL}, 2020.

\bibitem{lee2020context}
K.~Lee, Y.~Seo, S.~Lee, H.~Lee, and J.~Shin, ``Context-aware dynamics model for
  generalization in model-based reinforcement learning,'' in \emph{{ICML}},
  2020.

\bibitem{achterhold2021explore}
J.~Achterhold and J.~Stueckler, ``Explore the {C}ontext: Optimal data
  collection for context-conditional dynamics models,'' in \emph{{AISTATS}},
  2021.

\bibitem{garnelo2018neural}
M.~Garnelo, J.~Schwarz, D.~Rosenbaum, F.~Viola, D.~J. Rezende, S.~M.~A. Eslami,
  and Y.~W. Teh, ``Neural {P}rocesses,'' \emph{CoRR}, vol. abs/1807.01622,
  2018.

\bibitem{cho2014learning}
K.~Cho, B.~van Merrienboer, {\c{C}}.~G{\"{u}}l{\c{c}}ehre, D.~Bahdanau,
  F.~Bougares, H.~Schwenk, and Y.~Bengio, ``Learning phrase representations
  using {RNN} encoder-decoder for statistical machine translation,'' in
  \emph{{EMNLP}}, 2014.

\bibitem{le2018empirical}
T.~A. Le, H.~Kim, M.~Garnelo, D.~Rosenbaum, J.~Schwarz, and Y.~W. Teh,
  ``Empirical evaluation of neural process objectives,'' in \emph{Third
  Workshop on Bayesian Deep Learning}, 2018.

\bibitem{rubinstein1999cross}
R.~Rubinstein, ``The cross-entropy method for combinatorial and continuous
  optimization,'' \emph{Methodology and Computing in Applied Probability},
  1999.

\bibitem{pinneri2020sample}
C.~Pinneri, S.~Sawant, S.~Blaes, J.~Achterhold, J.~Stueckler,
  M.~Rol{\'{\i}}nek, and G.~Martius, ``Sample-efficient cross-entropy method
  for real-time planning,'' in \emph{CoRL}, 2020.

\end{thebibliography}
\bibliographystyle{IEEEtran}

\section*{CORRECTIONS}
In this version, we have added a link to a repository with source code for our approach at \url{https://github.com/EmbodiedVision/tradyn}. 
We have added a statement that we evaluate all models after a fixed number of 100k training steps (\cref{sec:tradyn:modeltraining}).
We have clarified that, for model training, we use a normalized velocity as observation ($v / 5$) (\cref{sec:tradyn:simenv_robdyn}). 
The ECMR version of this paper reports the velocity prediction error on the normalized velocity ($v / 5$). 
This version reports the velocity prediction error on the non-normalized velocity $v$ (\cref{fig:tradyn:unicycle_prediction}, center panel). 
The code to compute the numbers in \cref{fig:tradyn:unicycle_planning_gain} contained an implementation error, which caused only models of equal seeds (i.e., $i_1 = i_2$ in the caption of \cref{fig:tradyn:unicycle_planning_gain}) to be compared, but not across different seeds. 
We have corrected this for this version, by which the numbers have changed slightly (max.~abs.~difference in median ctrl.~energy: $0.004$). 
The significance statements and variant order w.r.t. median control energy ($+$T,$+$C $<$ $+$T,$-$C $<$ $-$T,$+$C $<$ $-$T,$-$C) remain unchanged. 
The updated code is available at \url{https://github.com/EmbodiedVision/tradyn}.
\end{document}